\documentclass[10pt,journal,compsoc]{IEEEtran}
\ifCLASSOPTIONcompsoc
  \usepackage[nocompress]{cite}
\else
  \usepackage{cite}
\fi
\ifCLASSINFOpdf
\else
\fi
\usepackage{algorithm}
\usepackage{algorithmic}
\usepackage{caption}
\usepackage[utf8]{inputenc} 
\usepackage[T1]{fontenc}    
\usepackage{hyperref}       
\usepackage{url}            
\usepackage{booktabs}       
\usepackage{amsfonts}       
\usepackage{nicefrac}       
\usepackage{microtype}      
\usepackage{graphicx}
\usepackage{subfigure}
\usepackage{amsmath}
\usepackage{amsmath}
\usepackage{color}
\usepackage{bm}
\usepackage{multirow}
\usepackage{float}

\begin{document}

\title{DAmageNet: A Universal Adversarial Dataset}

\author{Sizhe~Chen, Xiaolin~Huang$^*$, Zhengbao~He, Chengjin~Sun,
\IEEEcompsocitemizethanks{\IEEEcompsocthanksitem S.~Chen, X.~Huang, Z.~He and C.~Sun are with Institute of Image Processing and Pattern Recognition, Shanghai Jiao Tong University, also with the MOE Key Laboratory of System Control and Information Processing, 800 Dongchuan Road, Shanghai, 200240, P.R. China. (e-mails:\{csz729020210, xiaolinhuang, lstefanie, sunchengjin\}@sjtu.edu.cn)\protect\\
\IEEEcompsocthanksitem Corresponding author: Xiaolin Huang.}}

\IEEEtitleabstractindextext{
\begin{abstract}
It is now well known that deep neural networks (DNNs) are vulnerable to adversarial attack. Adversarial samples are similar to the clean ones, but are able to cheat the attacked DNN to produce incorrect predictions in high confidence. But most of the existing adversarial attacks have high success rate only when the information of the attacked DNN is well-known or could be estimated by massive queries. A promising way is to generate adversarial samples with high transferability. By this way, we generate 96020 transferable adversarial samples from original ones in ImageNet. The average difference, measured by root means squared deviation, is only around 3.8 on average. However, the adversarial samples are misclassified by various models with an error rate up to 90\%. Since the images are generated independently with the attacked DNNs, this is essentially zero-query adversarial attack. We call the dataset \emph{DAmageNet}, which is the first universal adversarial dataset that beats many models trained in ImageNet. By finding the drawbacks, DAmageNet could serve as a benchmark to study and improve robustness of DNNs. DAmageNet could be downloaded in \url{http://www.pami.sjtu.edu.cn/Show/56/122}
\end{abstract}
}

\maketitle
\IEEEdisplaynontitleabstractindextext
\IEEEpeerreviewmaketitle

\IEEEraisesectionheading{\section{Introduction}\label{sec:introduction}}
\IEEEPARstart{D}{eep} neural networks (DNNs) have grown into the mainstream methods in many fields, thus, their vulnerability has attacked attentions in the recent years. An obvious example is the existence of adversarial samples \cite{akhtar2018threat}, which are quite similar with the clean ones, but are able to cheat the DNN to produce incorrect prediction in high confidence. Various attack methods to craft adversarial samples have been proposed, including FGSM \cite{goodfellow2014explaining}, C\&W \cite{carlini2017towards}, PGD \cite{madry2017towards}, Type I \cite{tang2019adversarial} and so on. Generally speaking, when the victim network is exposed to the attacker, one can easily achieve efficient attack with very high successful rate.

Although white-box attacks can easily cheat DNNs, the current users actually do not worry about them, since it is almost impossible to get complete information including the structure and the parameters of the attacked DNNs. If the information is kept well, one has to use black-box attack, which can be roughly categorized into query approach \cite{cheng2019improving, ilyas2018prior, guo2019subspace} and transfer approach \cite{papernot2017practical, moosavi2017universal, dong2019evading}. The former one is to estimate the gradient by querying the attacked DNNs. However, until now, the existing query-based attack still needs massive queries, which can be easily detected by the defense systems.

Transfer approach attacks rely on the similarity between the attacked DNN and a surrogate model in the attacker's hands. It is expected that white-box attacks on the surrogate model can also invade the attacked DNN. Although there are some promising studies recently \cite{liu2016delving, xie2019improving, inkawhich2019feature}, the transfer performance is not satisfactory and high attack ratio could be reached only when the two DNNs are very similar, which however conflicts the aim of black-box attack.

\begin{figure}[htbp]
	\centering
	\includegraphics[width=\hsize]{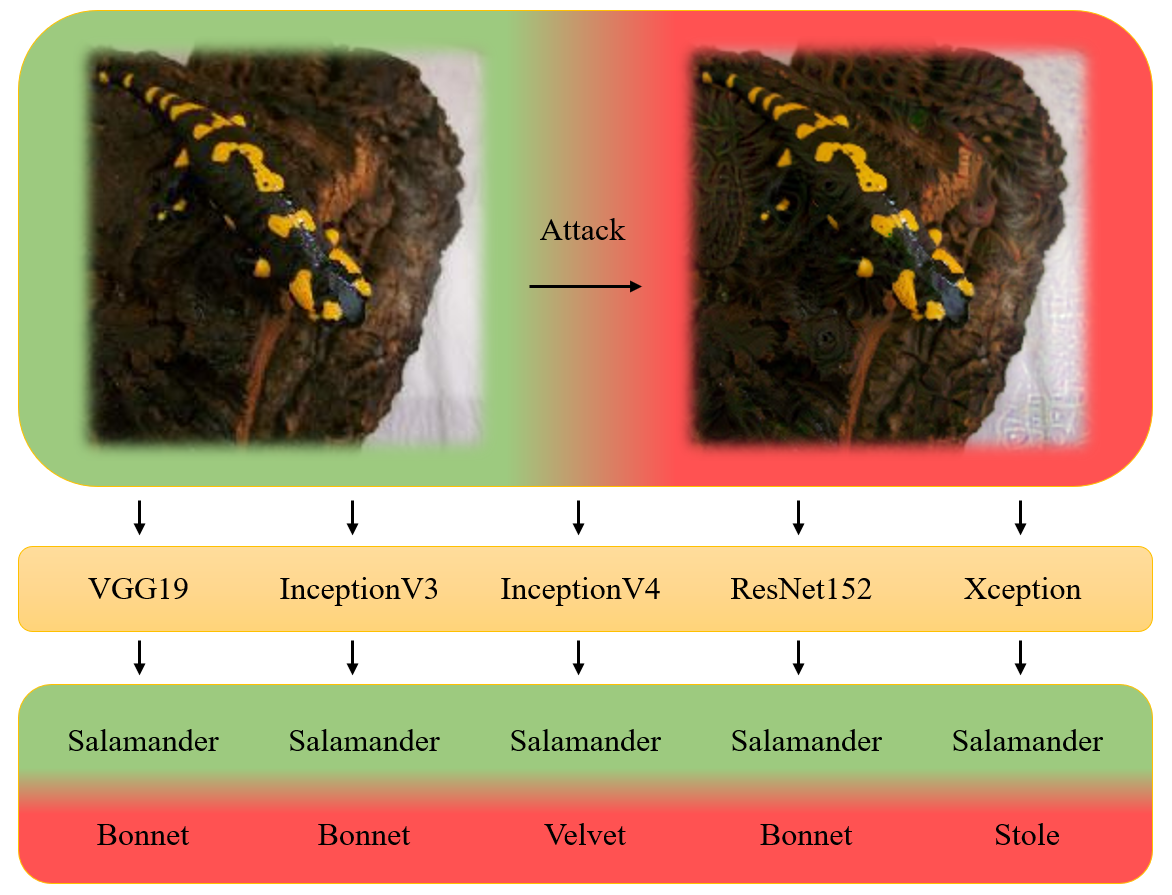}
	\caption{An adversarial sample in DAmageNet. The original sample (in ImageNet: image n01629819\_15314.JPEG, class No.25) is shown on the left. All well-trained DNNs (listed in the first row) correctly recognizes this image as a salamander. The right image is the generated adversarial sample in DAmageNet. The difference between the two images are slight, however, all the listed DNNs make incorrect prediction, as shown in the bottom row.}
	\label{intro}
\end{figure}

By attacking common vulnerabilities of DNNs, we successfully achieve zero-query black-box attack. In other words, we generate adversarial samples that can cheat many unknown DNNs. Fig. \ref{intro} illustrates an example. The original image is a "salamander" in ImageNet \cite{deng2009imagenet}. The adversarial one looks very similar to the original one but is misclassified by many DNNs well-trained in ImageNet.

From original images in ImageNet, we generate 96020 images that can cheat many well-trained DNNs. Similarly to the phenomenon shown above, these adversarial examples have very good black-box attack performance, i.e., it is in a zero-query mannor and make DNNs have up to 90\% error rate. We name this dataset as \emph{DAmageNet}.

To the best of our knowledge, DAmageNet is the first dataset that provides black-box adversarial attack images. Those images \emph{DAmage} many neural networks without any knowledge and any query. But the aim is not to really damage them, but to point out the weak parts of neural networks and thus those examples are valuable to improve the neural networks by means such as adversarial training \cite{ganin2016domain, shrivastava2017learning}, robustness certification \cite{sinha2017certifiable} and so on.

In the following, we will briefly introduce adversarial attack, especially black-box attack, and ImageNet in Section 2. Then DAmageNet will be described in Section 3 together with initial experiments. Section 4 ends the manuscript with a brief discussion.

\section{Related Work}
\subsection{Adversarial attack}
Adversarial attacks \cite{szegedy2013intriguing} could reveal the weakness of DNN by cheating it with adversarial samples, which differ from original ones with only slight perturbations. In humans' eyes, the adversarial sample does not have change from the original ones, but well-trained networks make false prediction in high confidence. The adversarial attack can be expressed as below.
\begin{eqnarray}\label{white}
\begin{split}
{\text { find }} & {\Delta \mathrm{x}} \\ {\text { s.t. }} & {f(\mathrm{x}) \neq f(\mathrm{x}+\Delta \mathrm{x})} \\ {} & {\|\Delta \mathrm{x}\| \leq \epsilon}
\end{split}
\end{eqnarray}
When training DNNs, one updates the weights of the network by the gradients with respective to the parameters to minimize the loss. While in adversarial attacks, one alters the image in the direction that increases the training loss. For this basic idea, there have been many variants on attacking space and crafting method.

For the attacking space, most of the existing directly impose attack in the image space, see, e.g., \cite{goodfellow2014explaining, moosavi2016deepfool, su2019one}. It is also reasonable to aim at latent feature vector space \cite{song2018constructing, tang2019adversarial} or the encoder/decoder \cite{baluja2017adversarial, han2019once}. Attack on feature spaces may have better semantic meanings and are harder to detect due to the perturbation unlike noise.

The methods that find the adversarial samples could be roughly categorized as gradient-based \cite{goodfellow2014explaining, madry2017towards} and optimization-based methods \cite{szegedy2013intriguing, carlini2017towards}. Gradient-based methods search on the gradient direction and the magnitude of perturbation is restricted in order to avoid distortion. Optimization-based method includes the magnitude restriction in the loss, i.e.,
\begin{eqnarray}\label{optimization}
{\mathop{\arg\min}_{\Delta \mathrm{x}}} & {L_{\mathrm{attack}} = L_{train} + \rho * d(x, x+\Delta \mathrm{x})}
\end{eqnarray}
where $\rho$ specifies the degree to restrict distortion, $d(\cdot,\cdot)$ could be $l_1$, $l_2$, $l_\infty$ norm or other distance metrics for different purposes. Adam \cite{kingma2014adam}, SGD \cite{bottou2010large} or other optimizer could be applied to optimize $L_{\mathrm{attack}}$ and achieve attack. $L_{\mathrm{attack}}$ could also vary to meet different requirements.

\subsection{Black-box attack}
When the attacked DNNs are totally known, the attacks mentioned above have high successful ratio. However, it is almost impossible to have access to the victim model in real scenarios and thus black-box attacks are required \cite{papernot2016transferability, brendel2017decision, ilyas2018black}. For black-box attacks, the attacker does not access of the attacked DNNs and the attacks relies on either query \cite{cheng2019improving, ilyas2018prior} or transferability \cite{papernot2016transferability, papernot2017practical}.

For query approach, the attacker adds slight perturbation to the input image and observes the reaction the victim model. By constant query, the gradient could be roughly estimated and then one can do attack in the way similar to white-box case. To choose the next pixel to alter for gradient estimation, Bayes optimization \cite{anonymous2020bayesopt}, evolutional algorithms \cite{laurent2019yet}, meta learning \cite{du2019query} etc could be considered. Since the practical DNNs are generally very complicated, good estimation of the gradients need massive number of queries, from which it follows easy detections.

For transfer approach, one conducts white-box attack in a well-designed surrogate model and expects that the adversarial samples remain aggressive to other models. The underlying assumption is that the distance between decision boundaries across different classes is significantly shorter than that across different models \cite{papernot2016transferability}. Although good attack ratio has been reported recently \cite{dong2019evading, inkawhich2019feature}, it heavily relies on good transferability of the surrogated and the attacked model, e.g., VGG16 and VGG19. Until now, adversarial samples which could beat many DNNs have not been reported and there is no publicly available dataset providing useful adversarial samples.

\subsection{Use of adversarial samples}
Since adversarial samples reveal the weak point of the DNNs, they could be used in many aspects, including robustness testing and training. For testing, every model is supposed to be put in non-cooperative environment and tested its performance in face of adversarial samples before being applied massively \cite{tsipras2018robustness, zhang2019theoretically}. For training, usually called adversarial training, the model is able to learn from its own weakness. When one re-trains a model, huge number of correct data is not as efficient as several data for which the current model made wrong prediction. Therefore, crafted adversarial samples are quite valuable for subsequent fine-tuning \cite{ganin2016domain, shrivastava2017learning}.

Since there is no efficient attack applicable to general DNNs, the current use of adversarial samples are model-dependent. That is the adversarial samples need to be generated for a given DNN. Our DAmageNet is the first data set providing adversarial samples for all DNNs. It could serve as a benchmark for robustness test and also could be directly used for adversarial training.

\subsection{ImageNet and its variants}
DAmageNet is obtained by modifying images from ImageNet. ImageNet is a universal dataset publicized by \cite{deng2009imagenet}. It contains images of 1000 classes, each has 1300 well-chosen samples. AlexNet \cite{krizhevsky2012imagenet} succeeds in emphasizing the superiority of deep learning by significantly outperforming rivals in ImageNet Large Scale Visual Recognition Challenge (ILSVRC). Recent years have seen the appearance of a lot of great models \cite{simonyan2014very, he2016deep, huang2017densely} and their improvements of model accuracy in ImageNet. Many interesting variants of ImageNet have been developed, including ImageNet-A \cite{hendrycks2019natural}, ObjectNet \cite{barbu2019objectnet}, ImageNet-C, and ImageNet-P \cite{hendrycks2019benchmarking}.

ImageNet-A contains real-world images in ImageNet classes, but they are able to mislead current classifiers to output false predictions. They are real images out of the manifold of ImageNet. The researchers select 200 typical classes from original 1000 classes and collect 7500 samples in total. Existing networks trained on ImageNet, even with robustness enhancements, cannot work well on them.

ObjectNet also contains natural images that pretrained models in ImageNet cannot distinguish. It controls that the object has random background, rotation and viewpoint. The models suffer from a performance drop up to 40-45\% in testing in ObjectNet. It contains 50000 images as the validation set in ImageNet.

ImageNet-C is produced from ImageNet validation images by adding 15 diverse corruptions. Each type of corruption has 5 levels from the lightest to the severest. It contains 50000 samples as in ImageNet validation set. It aims to simulate common image corruptions in real scenarios and their influence of DNN prediction. ImageNet-P is designed from ImageNet-C and differs from it in possessing additional perturbation sequences, which are not generated by adversarial attack but are common image transformations. The perturbations such as gaussian noise, motion blur, snow, brightness, translation, and rotation are sampled randomly and added to images repeated.

The datasets mentioned above are very valuable for test and improving the generalization capability, but DAmageNet is for robustness. In other words, samples in the above datasets are different from the samples in ImageNet and the low accuracy is due to lack of data. In DAmageNet, the samples are quite similar to the original ones in ImageNet and the low accuracy is due to the over-sensitivity of the DNNs. Other difference is that adversarial samples are relatively easy to obtain and the error is largely higher.

\section{DAmageNet}
\subsection{Description}
\begin{figure}[htbp]
	\centering
	\includegraphics[width=\hsize]{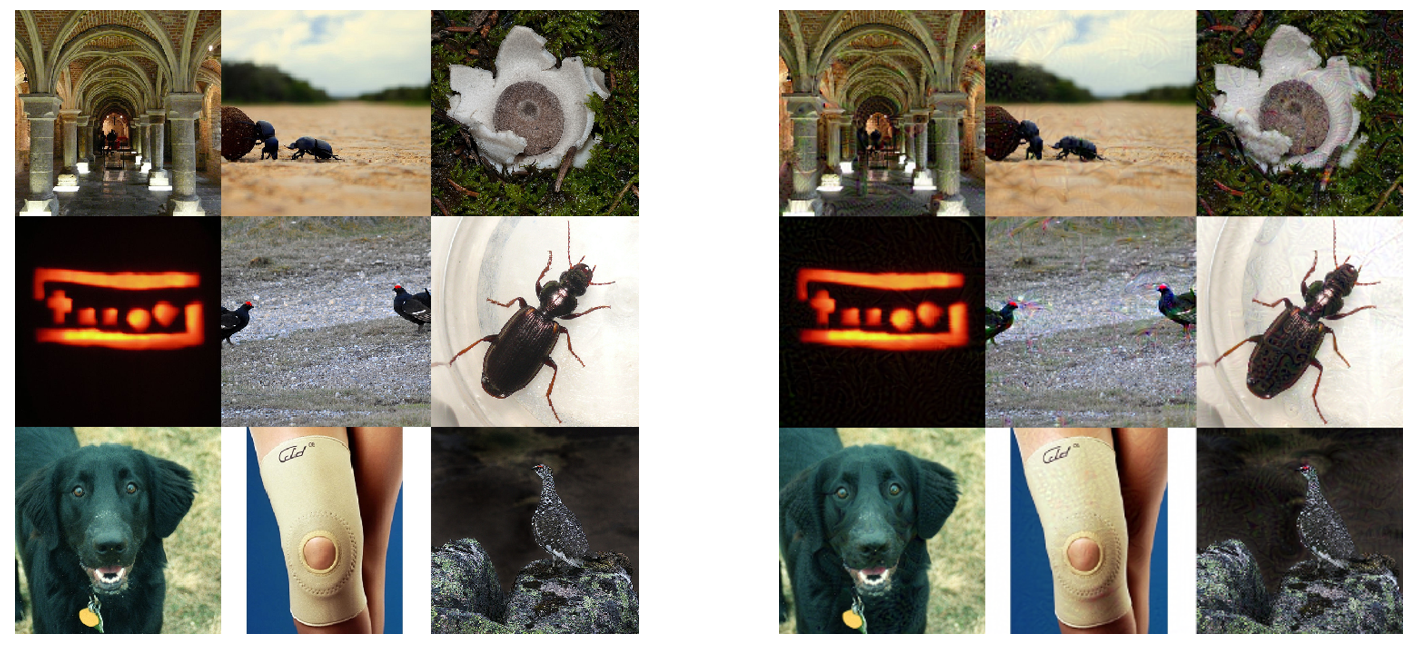}
	\caption{Samples in ImageNet and DAmageNet. The images on the left are original samples from ImageNet. The images on the right are adversarial samples from DAmageNet. One could observe that these images look similarly and human-beings have no problem to recognize them as the same class.
}
	\label{samples}
\end{figure}

DAmageNet contains 96020 adversarial samples in total and it could be downloaded from \url{http://www.pami.sjtu.edu.cn/Show/56/122}. The dataset contains 1000 folders with the correct labels as the folder name, which are in the order of ImageNet. In each of the folders, one can find around 100 adversarial samples. The file name is the index given to the original image in ImageNet, by that one can easily find the corresponding original image. The samples in DAmageNet are very similar to those in ImageNet training set and the root mean square deviation ($||\Delta x||_{2}$) is about 3.8. In Fig. \ref{samples}, we demonstrate image pairs in ImageNet and DAmageNet.

\subsection{Results}
DAmageNet is to provide adversarial samples applicable to many DNNs. Here, we use several pre-trained models in Keras Applications \cite{chollet2015keras} to recognize the images in DAmageNet, clean images in DAmageNet. Compared to the ImageNet results in original studies, the error rate (top-1) is reported Table \ref{result}. Notice that when generating DAmageNet, we do not access those attacked DNNs and the high error rate shows that samples in DAmageNet could do zero-query black-box attack. Thus, these samples could be used as a benchmark to measure the robustness and to improve DNNs.

\begin{table}[htbp]
	\caption{Error Rate (Top-1) in DAmageNet and ImageNet ({\small *\emph{For ImageNet, we only consider the images that generate DAmageNet, in oder to show the attack performance. For the error rates on the whole ImageNet, please refer to the literatures.} })}
	\centering
	\begin{tabular}{l|ccc}
    \toprule
		Network & DAmageNet & ImageNet\\
    \midrule
        VGG16 \cite{simonyan2014very} & 99.7\% & 12.6\% \\
        VGG19 \cite{simonyan2014very} & 100.0\% & 5.1\% \\
        ResNet50 \cite{he2016deep} & 92.5\% & 11.4\% \\
        ResNet101 \cite{he2016deep} & 84.6\% & 17.3\% \\
        ResNet152 \cite{he2016deep} & 81.8\% & 16.6\% \\
        NASNetMobile \cite{zoph2018learning} & 90.3\% & 13.2\% \\
        NASNetLarge \cite{zoph2018learning} & 100.0\% & 4.8\% \\
        InceptionV3 \cite{szegedy2016rethinking} & 96.7\% & 6.4\% \\
        InceptionV4 \cite{szegedy2017inception} & 100.0\% & 11.7\% \\
        Xception \cite{chollet2017xception} & 100.0\% & 8.8\% \\
        DenseNet121 \cite{huang2017densely} & 100.0\% & 15.2\% \\
        DenseNet169 \cite{huang2017densely} & 94.3\% & 10.8\% \\
        DenseNet201 \cite{huang2017densely} & 91.6\% & 9.5\% \\
        CondenseNet74-4 \cite{huang2018condensenet} & 95.5\% & 18.3\% \\
        CondenseNet74-8 \cite{huang2018condensenet} & 93.1\% & 22.5\% \\
    \bottomrule
	\end{tabular}
	\label{result}
\end{table}

\section{Conclusion}
We provide DAmageNet, a dataset containing universal adversarial samples. It is the first dataset that have samples with small perturbation, strong aggression, and then high cross-model attack ratio. It demonstrates that using zero-query adversarial attack to create generalized adversarial samples is possible, which is also a caution that without robustness enhancement (e.g., attack-against filter, attack-aware detector) DNNs are very vulnerable. DAmageNet provides benchmark from a different aspect to evaluate the robustness of DNN by elaborately-crafted adversarial samples and offers a helpful tool to further study transferability of attack. Since its transferability, the defense designed for DAmageNet may also have universality for many DNNs. Our algorithm to craft DAmageNet will be explained in detail soon in another paper.

\ifCLASSOPTIONcompsoc
  \section*{Acknowledgments}
\else
  \section*{Acknowledgment}
\fi
This work was partially supported by National Key Research Development Project (No. 2018AAA0100702), National Natural Science Foundation of China (No. 61977046), and 1000-Talent Plan (Young Program).
\ifCLASSOPTIONcaptionsoff
  \newpage
\fi

\bibliographystyle{IEEEtran}
\bibliography{Reference}

\end{document}